\documentclass[sigconf,natbib=true]{acmart}

\setcopyright{acmlicensed}
\copyrightyear{2026}
\acmYear{2026}
\acmDOI{XXXXXXX.XXXXXXX}
\acmConference[SIGIR '26]{the 49th International ACM SIGIR Conference on Research and Development in Information Retrieval}{April 20-24, 2026}{Melbourne, Australia}

\usepackage{libertine}
\usepackage{booktabs}
\usepackage{multirow}
\usepackage{pifont}
\usepackage{xspace}
\usepackage{enumitem}
\usepackage[table]{xcolor} 
\usepackage{minted}
\usepackage[most]{tcolorbox}

\definecolor{customblue}{rgb}{0.168, 0.364, 0.557}
\colorlet{framegray}{gray!3!white}
\newcommand{\dataset}{\textsc{Parse}\xspace}

\begin{document}

\title{\dataset: An Open-Domain Reasoning Question Answering Benchmark for Persian}

\author{Jamshid Mozafari}
\authornotemark[1]
\orcid{0000-0003-4850-9239}
\affiliation{%
  \institution{University of Innsbruck}
 \city{Innsbruck}
  \country{Austria}
  }
\email{jamshid.mozafari@uibk.ac.at}

\author{Seyed Parsa Mousavinasab}
\authornote{Equal contribution.}
\orcid{0009-0008-1652-4956}
\affiliation{%
  \institution{University of Innsbruck}
 \city{Innsbruck}
  \country{Austria}
  }
\email{seyed.mousavinasab@student.uibk.ac.at}

\author{Adam Jatowt}
\orcid{0000-0001-7235-0665}
\affiliation{%
  \institution{University of Innsbruck}
  \city{Innsbruck}
  \country{Austria}
  }
\email{adam.jatowt@uibk.ac.at}



\begin{abstract}
Reasoning-focused Question Answering (QA) has advanced rapidly with Large Language Models (LLMs), yet high-quality benchmarks for low-resource languages remain scarce. Persian, spoken by roughly 130 million people, lacks a comprehensive open-domain resource for evaluating reasoning-capable QA systems.
We introduce \dataset, the first open-domain Persian reasoning QA benchmark, containing 10{,}800 questions across Boolean, multiple-choice, and factoid formats, with diverse reasoning types, difficulty levels, and answer structures. The benchmark is built via a controlled LLM-based generation pipeline and validated through human evaluation.
We also ensure linguistic and factual quality through multi-stage filtering, annotation, and consistency checks.
We benchmark multilingual and Persian LLMs under multiple prompting strategies and show that Persian prompts and structured prompting (CoT for Boolean/multiple-choice; few-shot for factoid) improve performance. Fine-tuning further boosts results, especially for Persian-specialized models. These findings highlight how \dataset\ supports both fair comparison and practical model adaptation. 
\dataset\ fills a critical gap in Persian QA research and provides a strong foundation for developing and evaluating reasoning-capable LLMs in low-resource settings.
\end{abstract}






\maketitle

\section{Introduction}\label{s:introduction}
\begin{figure}[t]
  \centering
  \includegraphics[width=0.9\columnwidth]{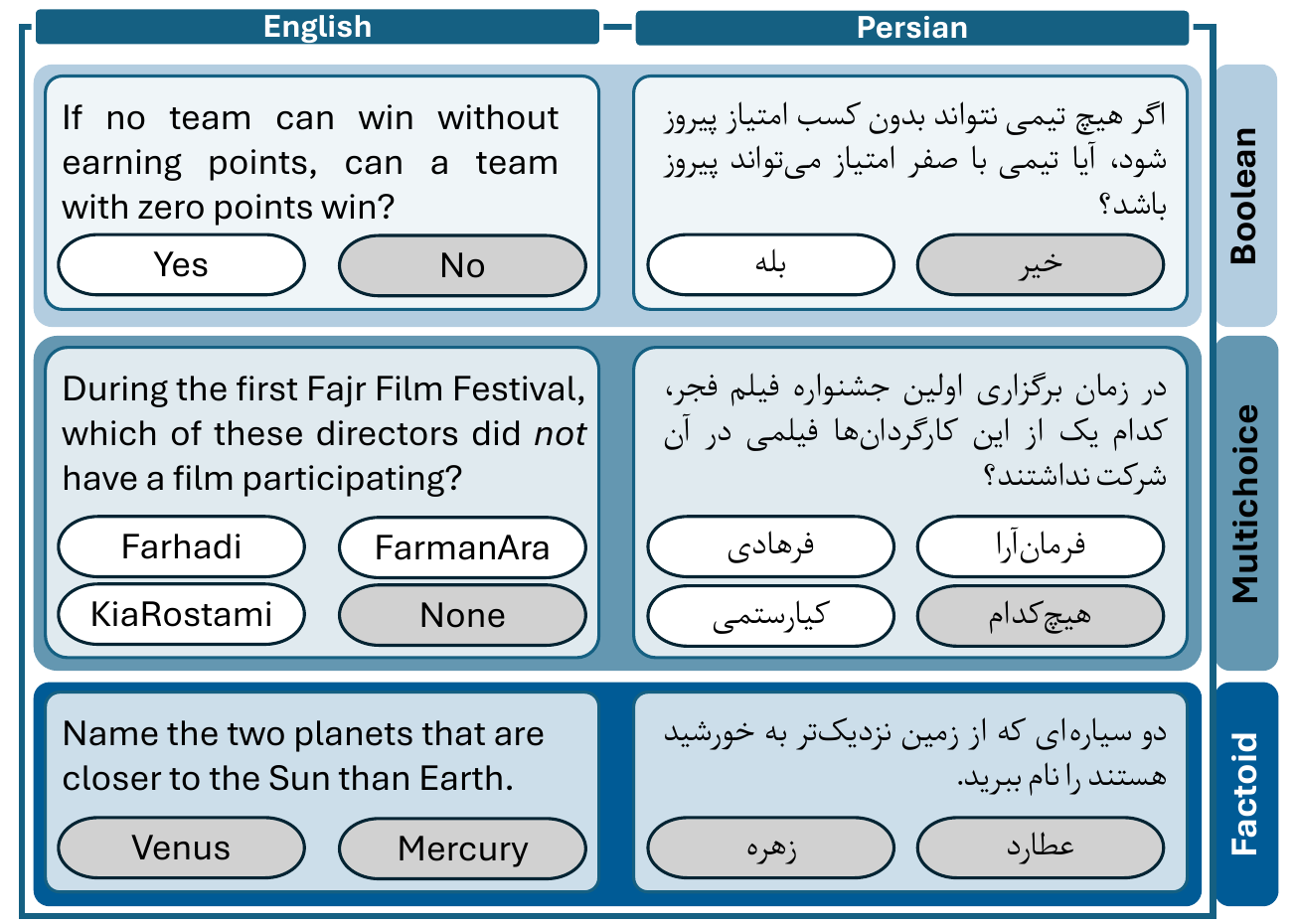}
  \caption{Sample questions from \dataset. The English column provides translations of the corresponding Persian questions. \textcolor{gray!80}{Gray-highlighted} options indicate the correct answers.}

  \label{fig:dataset_sample}
\end{figure}

\begin{table*}[t]
\centering
\caption{Comparison of major English (EN) and Persian (FA) QA datasets in terms of size, release year, difficulty-labels, and QA task types, including Boolean, multiple-choice, factoid, multi-hop, reasoning, multi-answer, and unanswerable questions.}
\label{tbl:persian_related_work}
\small
\begin{tabular}{l@{\hspace{30pt}} l l l c c c c c c c c}
\toprule
\textbf{Dataset} & \textbf{Lang} & \textbf{Size} & \textbf{Year} & \textbf{Diff-Labels} & \textbf{Bool} & \textbf{MCQ} & \textbf{Factoid} & \textbf{Multihop} & \textbf{Reasoning} & \textbf{Multi-Ans} & \textbf{Non-Ans} \\
\midrule
RACE~\cite{lai-etal-2017-race}                & EN & 100k  & 2017 &  &  & \checkmark & \checkmark &  & \checkmark &  &  \\
SQuAD2.0~\cite{rajpurkar-etal-2018-know}      & EN & 150k  & 2018 &  & &  & \checkmark &  &  &  & \checkmark \\
HotpotQA~\cite{yang-etal-2018-hotpotqa}      & EN & 113k  & 2018 & &  \checkmark &  & \checkmark & \checkmark & \checkmark &  &  \\
MultiRC~\cite{khashabi-etal-2018-looking}     & EN & 10k   & 2018 & &  & \checkmark & \checkmark & \checkmark & \checkmark & \checkmark &  \\
NaturalQuestions~\cite{kwiatkowski-etal-2019-natural} & EN & 307k  & 2019 & &  \checkmark &  & \checkmark &  &  &  & \checkmark \\
\midrule
\midrule
PerCQA~\cite{jamali-etal-2022-percqa}         & FA & 1k     & 2021 &  &  &  & \checkmark &  &  & \checkmark &  \\
PersianQA~\cite{PersianQA}                    & FA & 9k     & 2021 &  &  &  & \checkmark &  &  & \checkmark & \checkmark \\
PersianQuAD~\cite{9729745}                    & FA & 19.6k  & 2022 &  & &  & \checkmark &  &  &  &  \\
PQuAD~\cite{DARVISHI2023101486}               & FA & 80k    & 2023 & &  &  & \checkmark &  &  &  & \checkmark \\
IslamicPCQA~\cite{11075543}                   & FA & 12.3k  & 2023 & & \checkmark &  & \checkmark & \checkmark &  &  &  \\
PersianMedQA~\cite{kalahroodi2025persianmedqa} & FA & 20.8k & 2025 & &  & \checkmark & \checkmark &  & \checkmark &  &  \\
PersianMHQA~\cite{10.1145/3711826}            & FA & 7k     & 2025 & & \checkmark &  & \checkmark & \checkmark &  &  &  \\
\midrule
\textbf{\dataset (Ours)} & \textbf{FA} & \textbf{10.8k} & \textbf{2025} & \textbf{\checkmark} & \textbf{\checkmark} & \textbf{\checkmark} & \textbf{\checkmark} & \textbf{\checkmark} & \textbf{\checkmark} & \textbf{\checkmark} & \textbf{\checkmark} \\
\bottomrule
\end{tabular}
\end{table*}

Question Answering (QA) aims to provide accurate responses to user queries~\cite{pandya2021question}. With the advent of Large Language Models (LLMs)~\cite{grattafiori2024llama, team2025gemma, yang2025qwen3}, recent QA systems have progressed beyond traditional extractive settings to address more complex questions that require multi-step reasoning~\cite{patel-etal-2024-multi}. Such reasoning questions often require logical inference, conceptual linking, contextual understanding, and drawing conclusions, rather than simple pattern matching or direct extraction from text~\cite{plaat2024reasoning}.

LLMs have substantially reshaped the landscape of Natural Language Processing (NLP)~\cite{10.3389/frai.2023.1350306} and Information Retrieval (IR)~\cite{10.1145/3748304}. Their emergence has both raised performance expectations for existing tasks and enabled new research directions previously considered impractical. Notable topics include Retrieval-Augmented Generation (RAG)~\cite{10.5555/3495724.3496517}, hallucination detection~\cite{10.1145/3703155}, evaluation beyond accuracy~\cite{anonymous2024beyond}, instruction tuning~\cite{shengyu2023instruction}, multi-step reasoning~\cite{patel-etal-2024-multi}, and agent-based systems~\cite{luo2025large}.

Among these directions, reasoning-oriented QA~\cite{khashabi2019reasoning} has attracted significant attention. Prior work has explored a variety of methodologies~\cite{wu2024gendec, kim-etal-2024-learning} and datasets~\cite{khashabi-etal-2018-looking, zhang-etal-2023-crt}—primarily in English, with more recent efforts in languages such as Chinese~\cite{you2025benchmarking, E230044}. This growing body of work suggests that next-generation QA systems must exhibit strong reasoning capabilities, and that the ability of an LLM to answer reasoning questions is an important indicator of its overall capability~\cite{10973516}.

However, research on reasoning QA in low-resource languages remains limited, primarily due to the scarcity of available datasets and benchmarks. One notable example is Persian, a language with a long cultural history~\cite{Windfuhr2009TheIL} and approximately 130 million speakers worldwide\footnote{\url{https://en.wikipedia.org/wiki/Persian_language}}. To the best of our knowledge, there exists no open-domain benchmark designed to evaluate reasoning QA in Persian.

To address this gap, we introduce \dataset\footnote{\url{https://github.com/DataScienceUIBK/Parse}}, the first open-domain reasoning QA benchmark for Persian with general and cross-topic knowledge. \dataset contains 10,800 questions spanning diverse formats—including Boolean, multiple-choice, and factoid—and covering multiple reasoning categories such as multi-hop and complex inference questions, with single-answer, multi-answer, and unanswerable cases included. This breadth of coverage makes \dataset\ a comprehensive resource for evaluating LLM reasoning performance in Persian QA. Furthermore, given the scarcity of high-quality Persian LLMs~\cite{abbasi2023persianllama, rostami2024persianmind}, \dataset\ enables systematic benchmarking and comparison. Figure~\ref{fig:dataset_sample} presents representative examples from the benchmark.

In summary, our contributions are as follows:
\begin{enumerate}
\item We introduce \dataset, the first open-domain reasoning QA benchmark in Persian, comprising 10,800 diverse questions across multiple question and answer types.
\item We perform two human evaluation studies to validate benchmark quality.
\item We conduct comprehensive experiments using multilingual and Persian LLMs, demonstrating the utility of \dataset\ and analyzing the effects of fine-tuning.
\end{enumerate}

\section{Related Work}\label{s:related_work}
English question answering (QA) has seen the development of numerous large-scale benchmarks covering diverse task types and reasoning challenges. SQuAD2.0~\cite{rajpurkar-etal-2018-know} introduced more than 150k extractive questions, including unanswerable cases to evaluate abstention. NaturalQuestions (NQ)~\cite{kwiatkowski-etal-2019-natural} contains over 300k real-user queries from Google, supporting boolean and factoid QA as well as unanswerable questions. RACE~\cite{lai-etal-2017-race} offers 100k multiple-choice questions drawn from English examinations, designed to test reasoning. HotpotQA~\cite{yang-etal-2018-hotpotqa} includes 113k multi-hop questions requiring reasoning over multiple documents, while MultiRC~\cite{khashabi-etal-2018-looking} focuses on multi-sentence reading comprehension with multi-answer questions and explicit reasoning requirements. Collectively, these English benchmarks cover a comprehensive spectrum of QA task types, including boolean, multiple-choice, factoid, multi-hop, reasoning, multi-answer, and unanswerable questions.

In contrast, Persian QA resources~\cite{9443126, ijwr-parsquad, Kazemi2023, 3661286} remain narrow in scope, with each existing dataset covering only a subset of QA task dimensions. PersianQuAD~\cite{9729745} and PersianQA~\cite{PersianQA} provide factoid extractive QA, while PerCQA~\cite{jamali-etal-2022-percqa} focuses on community QA with multi-answer supervision. PQuAD~\cite{DARVISHI2023101486} expands factoid QA to 80k examples and introduces unanswerable questions, but does not include multi-hop or reasoning questions. More specialized efforts target specific domains or reasoning styles. PersianMedQA~\cite{kalahroodi2025persianmedqa} evaluates medical reasoning, whereas IslamicPCQA~\cite{11075543} and PersianMHQA~\cite{10.1145/3711826} focus on multi-hop QA. However, these benchmarks remain limited either by domain (e.g., medical) or by task type (e.g., multi-hop only). None jointly support the broad range of QA phenomena observed in English benchmarks. Table~\ref{tbl:persian_related_work} summarizes key differences across Persian datasets.

To the best of our knowledge, no existing Persian QA benchmark provides comprehensive open-domain coverage across boolean, multiple-choice, factoid, multi-hop, reasoning, multi-answer, and unanswerable question types, spanning general, cross-topic knowledge rather than a single specialized domain. We introduce \dataset, the first open-domain Persian reasoning QA benchmark that spans this full spectrum. By unifying major QA task types into a single resource, \dataset establishes a more challenging evaluation setting for Persian, enabling broader investigations into open-domain reasoning and bridging the gap with well-established English QA benchmarks.

\section{\dataset Benchmark}\label{s:dataset}
\begin{table*}[t]
\centering
\caption{Categorization of the \dataset benchmark by question type, reasoning dimension, and subtype, along with the number of questions.}
\label{tbl:dataset-statistics}

\setlength{\tabcolsep}{6pt}
\renewcommand{\arraystretch}{1.15}

\resizebox{\textwidth}{!}{%
\begin{tabular}{l r @{\hspace{40pt}} l r | l r @{\hspace{40pt}}l r | l r @{\hspace{40pt}}l r}
\toprule
\multicolumn{4}{c|}{\textbf{Boolean}} &
\multicolumn{4}{c|}{\textbf{Multi-choice}} &
\multicolumn{4}{c}{\textbf{Factoid}} \\
\cmidrule(lr){1-4}\cmidrule(lr){5-8}\cmidrule(lr){9-12}

\multicolumn{2}{c}{\textbf{Multihop}} &
\multicolumn{2}{c|}{\textbf{Reasoning}} &
\multicolumn{2}{c}{\textbf{Multihop}} &
\multicolumn{2}{c|}{\textbf{Reasoning}} &
\multicolumn{2}{c}{\textbf{Multihop}} &
\multicolumn{2}{c}{\textbf{Reasoning}} \\
\midrule

Simple         & 600 & Simple         & 600 &
Single-Ans   & 600 & Single-Ans   & 600 &
Simple         & 600 & Simple         & 600 \\

Negation       & 600 & Negation       & 600 &
Multi-Ans    & 600 & Multi-Ans    & 600 &
List-based      & 600 & List-based      & 600 \\

Comparative    & 600 & Comparative    & 600 &
Non-Ans & 600 & Non-Ans & 600 &
Non-Ans & 600 & Non-Ans & 600 \\
\midrule

\textbf{Total} & \textbf{1800} & \textbf{Total} & \textbf{1800} &
\textbf{Total} & \textbf{1800} & \textbf{Total} & \textbf{1800} &
\textbf{Total} & \textbf{1800} & \textbf{Total} & \textbf{1800} \\
\bottomrule
\end{tabular}%
}

\end{table*}

This section describes how we constructed \dataset, from prompt design to generation, assembly, and quality control. 

\paragraph{Task design and taxonomy.}
We began by defining three primary QA types—Boolean, Multiple-choice, and Factoid—and pairing each with two orthogonal approach dimensions: Multihop and Reasoning. Within these, we instantiated content subtypes reflecting common evaluation needs: for Boolean, \emph{simple}, \emph{negation}, and \emph{comparative}; for Multiple-choice, \emph{single-answer}, \emph{multi-answer}, and \emph{non-answerable}; and for Factoid, \emph{simple}, \emph{list-based}, and \emph{non-answerable}. Each question additionally receives a difficulty label (\emph{easy} /\emph{medium} /\emph{hard}) to support controlled evaluation. This cross-product yields 6 configurations per QA type\footnote{2 approaches $\times$ 3 subtypes = 6} and 18 configurations overall, providing broad coverage of form, reasoning requirements, and answer structure. Table~\ref{tbl:dataset-statistics} lists the taxonomy used to guide generation.

\paragraph{Prompting pipeline.}
We adopted an LLM-driven generation strategy using GPT-4o~\cite{achiam2023gpt}, followed by strict manual verification and filtering for each question. 
Rather than employing a single monolithic prompt, we designed \emph{configuration-specific} prompts that encode (i) the intended QA type and subtype, (ii) whether the item should require multihop and/or general reasoning, and (iii) a balanced difficulty schedule. Each prompt followed a consistent template with four components: a concise \emph{role} description anchoring the model’s behavior; a \emph{task} block specifying the target configuration; a set of \emph{requirements} that operationalize constraints (e.g., answer format, option count and ordering for Multiple-choice, use of Persian, realism, and topical diversity); and lightweight \emph{CSV-style output} instructions for downstream processing. This design makes the constraints explicit while keeping prompts short and reproducible. All prompt templates are publicly available in our GitHub repository\footnote{\url{https://github.com/DataScienceUIBK/Parse}}.

For each configuration family, we curated a dedicated prompt and generated batches of 30 questions per run (10 easy, 10 medium, 10 hard)\footnote{We chose 30 questions per batch because larger batches lowered output quality.}. We repeated the same prompt as needed to reach the desired sample size\footnote{We set the sample size to 600 per subtype.}. Throughout development, we monitored generation quality through spot checks, paying particular attention to Persian morphology and syntax; faithful realization of negation and comparative constructions; correct and balanced placement of answer options in Multiple-choice; and the use of genuine multihop evidence chains rather than single-hop recall.

\begin{figure}[t]
  \centering
  \includegraphics[width=0.7\columnwidth]{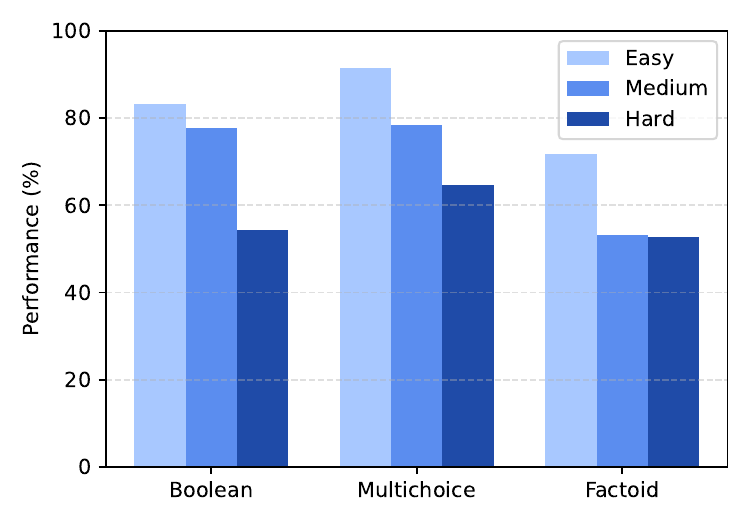}
  \caption{Human evaluation accuracy across difficulty levels (easy, medium, hard) for different question types.}

  \label{fig:dataset_difficulty}
\end{figure}
\begin{table}[t]
\caption{Average human ratings (1--5) for Ambiguity, Readability, and Correctness across four evaluation groups.}
\label{tbl:dataset_quality}
\resizebox{\columnwidth}{!}{%
\begin{tabular}{@{}l@{\hspace{70pt}}lll@{}}
\toprule
Group   & Ambiguity & Readability & Correctness \\ \midrule
Group 1 & 4.184     & 4.504       & 4.307       \\
Group 2 & 4.562     & 4.867       & 4.295       \\
Group 3 & 4.53      & 4.896       & 4.546       \\
Group 4 & 4.341     & 4.412       & 4.311       \\ 
\midrule
Average & \textbf{4.404}	    & \textbf{4.669}       & \textbf{4.389}       \\
\bottomrule
\end{tabular}%
}
\end{table}
\begin{table*}[t]
\caption{
Boolean QA performance under Zero-shot, Few-shot, and CoT prompting across English and Persian settings. 
Gray cells mark the higher of English vs.\ Persian for each model/subtype. 
Underlines indicate the strongest prompting strategy per model/language/subtype (ties excluded).
}
\label{tbl:boolean}
\resizebox{\textwidth}{!}{%
\begin{tabular}{@{}lccc|ccc||ccc|ccc@{}}
\toprule
\multicolumn{1}{c|}{\multirow{3}{*}{Model}} & \multicolumn{6}{c||}{English} & \multicolumn{6}{c}{Persian} \\ \cmidrule(l){2-13} 
\multicolumn{1}{c|}{} & \multicolumn{3}{c|}{Multihop} & \multicolumn{3}{c||}{Reasoning} & \multicolumn{3}{c|}{Multihop} & \multicolumn{3}{c}{Reasoning} \\
\multicolumn{1}{c|}{} & \multicolumn{1}{l}{Comparative} & \multicolumn{1}{l}{Negation} & \multicolumn{1}{l|}{Simple} & \multicolumn{1}{l}{Comparative} & \multicolumn{1}{l}{Negation} & \multicolumn{1}{l||}{Simple} & \multicolumn{1}{l}{Comparative} & \multicolumn{1}{l}{Negation} & \multicolumn{1}{l|}{Simple} & \multicolumn{1}{l}{Comparative} & \multicolumn{1}{l}{Negation} & \multicolumn{1}{l}{Simple} \\ \midrule
\multicolumn{13}{c}{Zero-Shot} \\ \midrule
\multicolumn{1}{l|}{Qwen 2.5 7b} & 0.49 & 0.57 & 0.78 & 0.71 & \cellcolor{gray!30}0.83 & 0.82 & \cellcolor{gray!30}0.53 & \cellcolor{gray!30}0.63 & \cellcolor{gray!30}0.79 & \cellcolor{gray!30}0.77 & 0.82 & \cellcolor{gray!30}0.83 \\
\multicolumn{1}{l|}{LLaMA 3 8B} & 0.54 & 0.69 & 0.77 & \cellcolor{gray!30}0.81 & \cellcolor{gray!30}0.65 & 0.69 & \cellcolor{gray!30}\underline{0.59} & \cellcolor{gray!30}\underline{0.78} & \cellcolor{gray!30}0.78 & 0.70 & 0.48 & \cellcolor{gray!30}0.71 \\
\multicolumn{1}{l|}{Mistral 24B} & 0.52 & 0.73 & 0.89 & 0.72 & \cellcolor{gray!30}0.74 & 0.85 & \cellcolor{gray!30}\underline{0.56} & \cellcolor{gray!30}\underline{0.93} & \cellcolor{gray!30}\underline{0.93} & \cellcolor{gray!30}0.84 & 0.71 & \cellcolor{gray!30}0.87 \\
\multicolumn{1}{l|}{Gemma 2 27B} & 0.57 & \cellcolor{gray!30}\underline{0.94} & \cellcolor{gray!30}0.94 & 0.86 & \cellcolor{gray!30}0.82 & \cellcolor{gray!30}0.92 & \cellcolor{gray!30}0.58 & \underline{0.93} & 0.93 & 0.86 & 0.79 & 0.91 \\
\multicolumn{1}{l|}{LLaMA 3 70B} & \cellcolor{gray!30}0.58 & 0.95 & 0.94 & 0.89 & 0.82 & 0.92 & 0.57 & \cellcolor{gray!30}0.96 & \cellcolor{gray!30}0.95 & \underline{0.89} & 0.82 & 0.92 \\
\multicolumn{1}{l|}{Qwen 2.5 72b} & 0.57 & 0.86 & 0.95 & 0.87 & \cellcolor{gray!30}0.82 & \cellcolor{gray!30}0.95 & 0.57 & \cellcolor{gray!30}0.88 & 0.95 & 0.87 & 0.81 & 0.94 \\ \midrule
\multicolumn{13}{c}{Few-Shot} \\ \midrule
\multicolumn{1}{l|}{Qwen 2.5 7b} & \underline{0.58} & \underline{0.64} & 0.84 & 0.75 & \cellcolor{gray!30}0.81 & \cellcolor{gray!30}0.84 & \cellcolor{gray!30}\underline{0.60} & \cellcolor{gray!30}\underline{0.85} & \cellcolor{gray!30}\underline{0.86} & \cellcolor{gray!30}0.81 & 0.67 & 0.81 \\
\multicolumn{1}{l|}{LLaMA 3 8B} & \cellcolor{gray!30}0.52 & \cellcolor{gray!30}\underline{0.87} & \cellcolor{gray!30}0.76 & \cellcolor{gray!30}\underline{0.82} & \cellcolor{gray!30}0.62 & \cellcolor{gray!30}0.76 & 0.45 & 0.55 & 0.56 & 0.51 & 0.57 & 0.63 \\
\multicolumn{1}{l|}{Mistral 24B} & 0.52 & 0.54 & 0.58 & 0.56 & 0.58 & 0.60 & \cellcolor{gray!30}0.54 & \cellcolor{gray!30}0.82 & \cellcolor{gray!30}0.82 & \cellcolor{gray!30}0.74 & \cellcolor{gray!30}\underline{0.73} & \cellcolor{gray!30}0.80 \\
\multicolumn{1}{l|}{Gemma 2 27B} & \underline{0.58} & 0.85 & 0.94 & 0.86 & \underline{0.89} & \cellcolor{gray!30}0.93 & 0.58 & \cellcolor{gray!30}0.88 & \cellcolor{gray!30}0.95 & 0.86 & \underline{0.89} & 0.92 \\
\multicolumn{1}{l|}{LLaMA 3 70B} & 0.58 & 0.95 & 0.94 & \cellcolor{gray!30}0.89 & \cellcolor{gray!30}\underline{0.90} & 0.94 & \underline{0.58} & \cellcolor{gray!30}0.96 & 0.94 & 0.87 & \underline{0.89} & 0.94 \\
\multicolumn{1}{l|}{Qwen 2.5 72b} & \cellcolor{gray!30}\underline{0.58} & \underline{0.89} & \underline{0.96} & 0.87 & \underline{0.90} & 0.95 & 0.57 & \cellcolor{gray!30}\underline{0.94} & \cellcolor{gray!30}\underline{0.97} & \cellcolor{gray!30}0.88 & \underline{0.90} & 0.95 \\ \midrule
\multicolumn{13}{c}{Chain-Of-Thought} \\ \midrule
\multicolumn{1}{l|}{Qwen 2.5 7b} & \cellcolor{gray!30}0.57 & \cellcolor{gray!30}0.63 & \cellcolor{gray!30}0.84 & \underline{0.82} & \underline{0.84} & \cellcolor{gray!30}\underline{0.89} & 0.56 & 0.62 & 0.81 & \cellcolor{gray!30}\underline{0.83} & \underline{0.84} & \underline{0.86} \\
\multicolumn{1}{l|}{LLaMA 3 8B} & \underline{0.55} & 0.47 & \cellcolor{gray!30}\underline{0.83} & \cellcolor{gray!30}0.80 & \cellcolor{gray!30}\underline{0.71} & \underline{0.81} & \cellcolor{gray!30}0.57 & \cellcolor{gray!30}0.62 & \underline{0.82} & \underline{0.78} & \underline{0.69} & \cellcolor{gray!30}\underline{0.83} \\
\multicolumn{1}{l|}{Mistral 24B} & \underline{0.54} & \underline{0.76} & \underline{0.92} & \underline{0.85} & \cellcolor{gray!30}\underline{0.75} & \cellcolor{gray!30}\underline{0.93} & \cellcolor{gray!30}0.55 & \cellcolor{gray!30}0.89 & 0.92 & \cellcolor{gray!30}\underline{0.86} & 0.69 & \underline{0.92} \\
\multicolumn{1}{l|}{Gemma 2 27B} & 0.56 & 0.87 & \cellcolor{gray!30}\underline{0.96} & \cellcolor{gray!30}0.86 & 0.85 & 0.93 & \cellcolor{gray!30}0.58 & \cellcolor{gray!30}0.90 & 0.95 & 0.85 & \cellcolor{gray!30}0.88 & \cellcolor{gray!30}\underline{0.94} \\
\multicolumn{1}{l|}{LLaMA 3 70B} & 0.57 & \cellcolor{gray!30}0.81 & 0.94 & \cellcolor{gray!30}0.88 & 0.85 & 0.94 & 0.57 & 0.80 & \cellcolor{gray!30}0.95 & 0.87 & \cellcolor{gray!30}0.87 & 0.94 \\
\multicolumn{1}{l|}{Qwen 2.5 72b} & 0.57 & 0.80 & 0.95 & 0.87 & 0.83 & 0.93 & \cellcolor{gray!30}\underline{0.58} & \cellcolor{gray!30}0.82 & 0.95 & \cellcolor{gray!30}\underline{0.89} & \cellcolor{gray!30}0.84 & \cellcolor{gray!30}\underline{0.96} \\
\bottomrule
\end{tabular}%
}
\end{table*}

\paragraph{Assembly and normalization.}
All runs were exported as CSV and concatenated per configuration. We then converted the data into a normalized JSON schema to enable programmatic validation. Each question receives a globally unique identifier, along with configuration-specific fields:
\texttt{question};
\texttt{answer} (a string for single-answer types, a list for list-based or multi-answer types, and a boolean for yes/no questions);
\texttt{options} for Multiple-choice items (exactly four entries);
and \texttt{difficulty} (easy, medium, or hard).

\paragraph{Quality control and balancing.}
We applied multiple passes of quality control.
First, we enforced structural validity: no missing fields (except for Factoid \emph{non-answerable}, which uses a conventional “None” placeholder), correct option cardinality for Multiple-choice, and valid difficulty labels.
Second, we removed exact duplicates within and across runs, and enforced option-level de-duplication in Multiple-choice items.
Third, we checked configuration-specific semantics: Boolean items must be answerable with “Yes/No” (Persian equivalents) and align with the proposition; Multiple-choice \emph{single-answer} must have exactly one valid answer and never “None”; \emph{multi-answer} must have two to four valid answers; and \emph{non-answerable} must include “None” as the only correct option. Factoid \emph{simple} and \emph{non-answerable} must contain exactly one answer string (with \emph{non-answerable} set to “None”), while \emph{list-based} must contain two to five gold responses.

Finally, we performed targeted checks for difficulty calibration and linguistic robustness. We ensured that \emph{hard} items derive their difficulty from reasoning depth, compositionality, or subtle semantic contrasts rather than convoluted grammar or obscure vocabulary; items failing this principle were rejected. We also screened for fluent, culturally natural Persian, favoring idiomatic usage and rejecting literal translations that could bias interpretation. These principles were embedded directly into prompt design and reinforced via post-hoc filtering to improve validity for both human and automatic evaluation.

After validation, we obtained uniform coverage: each of the 18 configuration families contains exactly 600 questions, evenly split by difficulty (200 easy / 200 medium / 200 hard), yielding \textbf{10{,}800} QA pairs in total. The finalized release preserves the taxonomy shown in Table~\ref{tbl:dataset-statistics}, enabling targeted ablations while supporting end-to-end evaluation across the full spectrum of QA phenomena in Persian.

\section{Human Evaluation}\label{s:human_evaluation}
Because relying solely on LLM-generated outputs risks systematic errors, we conducted human evaluation to verify the linguistic and factual quality of the generated QA pairs and to confirm the correctness of assigned difficulty labels. We performed two complementary evaluations: \textit{Quality Evaluation} and \textit{Difficulty Evaluation}.

\begin{table*}[]
\caption{
Multi-choice QA performance under Zero-shot, Few-shot, and CoT prompting across English and Persian settings. 
Gray cells mark the higher of English vs.\ Persian for each model/subtype. 
Underlines indicate the strongest prompting strategy per model/language/subtype (ties excluded).
}
\label{tbl:multichoice}
\resizebox{\textwidth}{!}{%
\begin{tabular}{@{}lccc|ccc||ccc|ccc@{}}
\toprule
\multicolumn{1}{c|}{\multirow{3}{*}{Model}} & \multicolumn{6}{c||}{English} & \multicolumn{6}{c}{Persian} \\ \cmidrule(l){2-13}
\multicolumn{1}{c|}{} & \multicolumn{3}{c|}{Multihop} & \multicolumn{3}{c||}{Reasoning} & \multicolumn{3}{c|}{Multihop} & \multicolumn{3}{c}{Reasoning} \\
\multicolumn{1}{c|}{} & Multi-Ans & Non-Ans & \multicolumn{1}{l|}{Single-Ans} & Multi-Ans & Non-Ans & \multicolumn{1}{l||}{Single-Ans} & Multi-Ans & Non-Ans & \multicolumn{1}{l|}{Single-Ans} & Multi-Ans & Non-Ans & Single-Ans \\ \midrule
\multicolumn{13}{c}{Zero-Shot} \\ \midrule
\multicolumn{1}{l|}{Qwen 2.5 7b} & \cellcolor{gray!30}\underline{0.83} & \cellcolor{gray!30}\underline{0.33} & 0.66 & \cellcolor{gray!30}0.78 & 0.48 & \cellcolor{gray!30}0.71 & 0.78 & \underline{0.29} & 0.66 & 0.77 & 0.48 & 0.69 \\
\multicolumn{1}{l|}{LLaMA 3 8B} & 0.76 & \cellcolor{gray!30}\underline{0.64} & 0.70 & 0.63 & \cellcolor{gray!30}\underline{0.74} & 0.48 & \cellcolor{gray!30}0.78 & \underline{0.45} & \cellcolor{gray!30}\underline{0.71} & \cellcolor{gray!30}0.65 & \underline{0.56} & \cellcolor{gray!30}0.49 \\
\multicolumn{1}{l|}{Mistral 24B} & 0.90 & 0.20 & 0.85 & 0.82 & \underline{0.46} & 0.80 & \underline{0.90} & 0.20 & 0.85 & 0.82 & \cellcolor{gray!30}\underline{0.52} & \cellcolor{gray!30}0.82 \\
\multicolumn{1}{l|}{Gemma 2 27B} & 0.92 & \cellcolor{gray!30}0.09 & \cellcolor{gray!30}0.89 & \cellcolor{gray!30}0.85 & 0.42 & 0.82 & 0.92 & 0.07 & 0.88 & 0.84 & 0.42 & 0.82 \\
\multicolumn{1}{l|}{LLaMA 3 70B} & \underline{0.95} & 0.05 & 0.91 & 0.86 & 0.33 & \cellcolor{gray!30}0.84 & \underline{0.95} & \cellcolor{gray!30}0.08 & 0.91 & 0.86 & \cellcolor{gray!30}0.40 & 0.83 \\
\multicolumn{1}{l|}{Qwen 2.5 72b} & 0.94 & 0.33 & \cellcolor{gray!30}0.90 & 0.88 & \underline{0.70} & 0.87 & 0.94 & \cellcolor{gray!30}0.38 & 0.88 & 0.88 & \cellcolor{gray!30}\underline{0.71} & 0.87 \\ \midrule
\multicolumn{13}{c}{Few-Shot} \\ \midrule
\multicolumn{1}{l|}{Qwen 2.5 7b} & 0.81 & 0.27 & \cellcolor{gray!30}0.73 & 0.76 & \underline{0.51} & \cellcolor{gray!30}0.76 & \cellcolor{gray!30}\underline{0.83} & \cellcolor{gray!30}0.28 & \underline{0.71} & \cellcolor{gray!30}0.77 & \cellcolor{gray!30}\underline{0.53} & 0.75 \\
\multicolumn{1}{l|}{LLaMA 3 8B} & 0.66 & 0.12 & 0.67 & 0.53 & 0.18 & 0.40 & \cellcolor{gray!30}0.79 & \cellcolor{gray!30}0.15 & \cellcolor{gray!30}0.70 & \cellcolor{gray!30}0.65 & \cellcolor{gray!30}0.24 & \cellcolor{gray!30}0.56 \\
\multicolumn{1}{l|}{Mistral 24B} & 0.43 & \underline{0.26} & 0.27 & 0.41 & 0.24 & 0.27 & \cellcolor{gray!30}0.46 & \cellcolor{gray!30}0.30 & \cellcolor{gray!30}0.28 & \cellcolor{gray!30}0.44 & \cellcolor{gray!30}0.27 & \cellcolor{gray!30}0.30 \\
\multicolumn{1}{l|}{Gemma 2 27B} & \cellcolor{gray!30}\underline{0.93} & \cellcolor{gray!30}\underline{0.29} & \underline{0.90} & 0.83 & \cellcolor{gray!30}\underline{0.60} & \cellcolor{gray!30}0.82 & 0.92 & \underline{0.26} & \underline{0.90} & \cellcolor{gray!30}0.84 & \underline{0.59} & 0.81 \\
\multicolumn{1}{l|}{LLaMA 3 70B} & 0.93 & \underline{0.25} & 0.91 & 0.83 & \underline{0.50} & \cellcolor{gray!30}0.86 & 0.93 & \cellcolor{gray!30}\underline{0.32} & \cellcolor{gray!30}\underline{0.92} & 0.83 & \cellcolor{gray!30}\underline{0.53} & 0.85 \\
\multicolumn{1}{l|}{Qwen 2.5 72b} & 0.94 & \cellcolor{gray!30}\underline{0.43} & \underline{0.91} & 0.88 & \cellcolor{gray!30}0.69 & 0.91 & 0.94 & \underline{0.40} & \underline{0.91} & \cellcolor{gray!30}0.89 & 0.66 & 0.91 \\ \midrule
\multicolumn{13}{c}{Chain-Of-Thought} \\ \midrule
\multicolumn{1}{l|}{Qwen 2.5 7b} & \cellcolor{gray!30}0.81 & 0.21 & \cellcolor{gray!30}\underline{0.74} & \cellcolor{gray!30}\underline{0.79} & 0.35 & \cellcolor{gray!30}\underline{0.88} & 0.76 & \cellcolor{gray!30}0.28 & 0.67 & 0.77 & \cellcolor{gray!30}0.39 & \underline{0.82} \\
\multicolumn{1}{l|}{LLaMA 3 8B} & \underline{0.78} & \cellcolor{gray!30}0.44 & \cellcolor{gray!30}0.70 & \cellcolor{gray!30}\underline{0.70} & \cellcolor{gray!30}0.59 & \underline{0.63} & \cellcolor{gray!30}\underline{0.80} & 0.36 & 0.65 & \underline{0.67} & 0.52 & \cellcolor{gray!30}\underline{0.64} \\
\multicolumn{1}{l|}{Mistral 24B} & \cellcolor{gray!30}0.90 & 0.25 & 0.85 & \underline{0.85} & 0.43 & \underline{0.92} & 0.89 & \cellcolor{gray!30}\underline{0.37} & 0.85 & \cellcolor{gray!30}\underline{0.86} & \cellcolor{gray!30}0.46 & \underline{0.92} \\
\multicolumn{1}{l|}{Gemma 2 27B} & 0.92 & 0.17 & \cellcolor{gray!30}0.89 & \cellcolor{gray!30}\underline{0.86} & 0.42 & \cellcolor{gray!30}\underline{0.94} & 0.92 & \cellcolor{gray!30}0.23 & 0.88 & \underline{0.85} & \cellcolor{gray!30}0.51 & \underline{0.91} \\
\multicolumn{1}{l|}{LLaMA 3 70B} & \cellcolor{gray!30}0.93 & 0.23 & 0.91 & \cellcolor{gray!30}\underline{0.88} & 0.44 & \underline{0.95} & 0.92 & \cellcolor{gray!30}0.28 & 0.91 & \underline{0.87} & \cellcolor{gray!30}0.48 & \underline{0.95} \\
\multicolumn{1}{l|}{Qwen 2.5 72b} & 0.92 & 0.20 & \cellcolor{gray!30}0.90 & \underline{0.89} & 0.39 & \underline{0.93} & 0.92 & \cellcolor{gray!30}0.27 & 0.88 & \cellcolor{gray!30}\underline{0.90} & \cellcolor{gray!30}0.43 & \underline{0.93} \\
\bottomrule
\end{tabular}%
}
\end{table*}

\paragraph{Quality evaluation.}

We assessed Ambiguity, Readability, and Correctness on a 1--5 scale.
From the full set of 10{,}800~QA pairs (54 configurations), we sampled 5~items per configuration, yielding 270~items per group. 
Four non-overlapping groups (1{,}080 items total, 10\% of the benchmark) were created; each group was independently evaluated by three native Persian speakers (12~annotators total), enabling majority agreement.

The annotators included 8~men and 4~women, spanning diverse educational backgrounds---Undergraduate (2~men, 1~woman), Bachelor's (4~men, 2~women), and Master's (2~men, 1~woman)---with ages ranging from 22~to~57.
Annotators used a lightweight web interface and rated each QA pair according to the three criteria. 
The annotation interface is available in our GitHub repository.

Table~\ref{tbl:dataset_quality} shows the averaged scores across all annotators. All three metrics scored above 4 on average, indicating (i)~clear and unambiguous questions, (ii)~grammatically fluent Persian, and (iii)~strong factual correctness.

\paragraph{Difficulty evaluation.}

To validate the difficulty labels (easy/medium/hard), three annotators evaluated a subset of 270 QA pairs. Each annotator answered five questions per configuration without being told the original difficulty label. They were informed only of the question type, but not the subtype, preventing potential bias and enabling a fairer assessment of question complexity.

Figure~\ref{fig:dataset_difficulty} shows that, across all question types, annotator performance is highest on easy questions, followed by medium and then hard questions. This alignment between human judgments and assigned difficulty labels supports the validity of the benchmark’s difficulty structure.

\section{Experimental Setup}\label{s:experimental_setup}
In our experiments, we evaluate six general-purpose LLMs---Qwen-2.5 (7B, 72B)~\cite{yang2025qwen3}, 
LLaMA-3 (8B, 70B)~\cite{grattafiori2024llama}, 
Mistral-24B~\cite{jiang2023mistral}, 
and Gemma-2 (27B)~\cite{team2025gemma}---covering diverse model families and sizes to minimize architecture- and scale-related bias.
We additionally include \textit{Dorna}\footnote{\url{https://huggingface.co/PartAI/Dorna-Llama3-8B-Instruct}}, 
a Persian-specific model trained on Persian corpora, to assess the impact of language specialization.

For evaluation, we use \textbf{Accuracy} for Boolean questions. 
For multiple-choice questions, we use \textbf{Accuracy} for single-answer and non-answerable variants, 
and \textbf{Jaccard} for multi-answer cases.
For factoid questions, we report \textbf{Accuracy} for non-answerable instances, 
\textbf{Contains} for simple string-matching answers, 
and \textbf{Jaccard} for list-style answers.

All experiments---including instruction tuning---were conducted on the Together~AI\footnote{\url{https://www.together.ai/}} 
platform with a decoding temperature of~0.7 on three different prompt strategies including Zero-Shot, Few-Shot, and Chain-of-Thought.
Prompts used for inference, grouped by question type, are available in our GitHub repository.

\section{Experiments}\label{s:experiments}
\begin{figure*}[t]
  \centering
  \includegraphics[width=\textwidth]{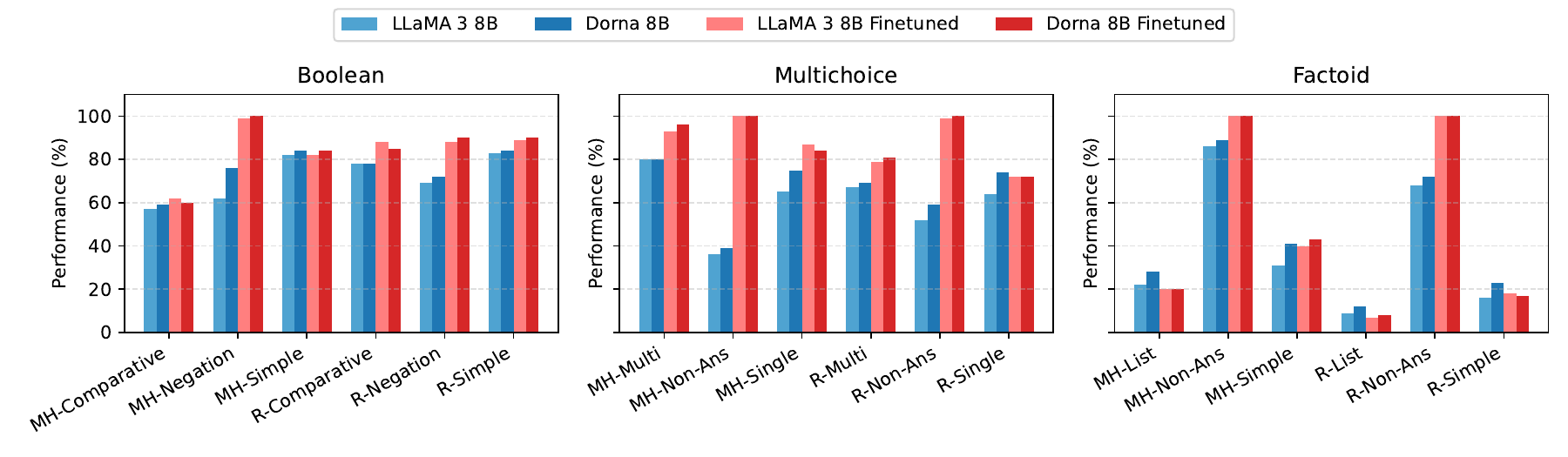}
  \caption{Performance of LLaMA~3~8B and Dorna before and after fine-tuning on \dataset. 
The figure reports evaluation scores on the 2{,}160-item test set sampled across all configurations.}

  \label{fig:finetune_results}
\end{figure*}

\begin{table*}[]
\caption{Factoid QA performance under Zero-shot, Few-shot, and CoT prompting across English and Persian settings. 
Gray cells mark the higher of English vs.\ Persian for each model/subtype. 
Underlines indicate the strongest prompting strategy per model/language/subtype (ties excluded).}
\label{tbl:factoid}
\resizebox{\textwidth}{!}{%
\begin{tabular}{@{}lccc|ccc||ccc|ccc@{}}
\toprule
\multicolumn{1}{c|}{\multirow{3}{*}{Model}} & \multicolumn{6}{c||}{English} & \multicolumn{6}{c}{Persian} \\ \cmidrule(l){2-13}
\multicolumn{1}{c|}{} & \multicolumn{3}{c|}{Multihop} & \multicolumn{3}{c||}{Reasoning} & \multicolumn{3}{c|}{Multihop} & \multicolumn{3}{c}{Reasoning} \\
\multicolumn{1}{c|}{} & List-based & Non-Ans & \multicolumn{1}{l|}{Simple} & List-based & Non-Ans & \multicolumn{1}{l||}{Simple} & List-based & Non-Ans & \multicolumn{1}{l|}{Simple} & List-based & Non-Ans & Simple \\ \midrule
\multicolumn{13}{c}{Zero-Shot} \\ \midrule
\multicolumn{1}{l|}{Qwen 2.5 7b} & 0.16 & \cellcolor{gray!30}0.82 & 0.22 & \cellcolor{gray!30}0.09 & 0.46 & 0.19 & 0.16 & 0.80 & \cellcolor{gray!30}0.23 & 0.08 & \cellcolor{gray!30}0.48 & \cellcolor{gray!30}0.22 \\
\multicolumn{1}{l|}{LLaMA 3 8B} & \cellcolor{gray!30}0.24 & 0.01 & 0.27 & 0.08 & 0.01 & \cellcolor{gray!30}0.18 & 0.23 & \cellcolor{gray!30}0.33 & 0.27 & 0.08 & \cellcolor{gray!30}0.27 & 0.17 \\
\multicolumn{1}{l|}{Mistral 24B} & 0.38 & 0.79 & 0.57 & 0.17 & 0.70 & 0.29 & \cellcolor{gray!30}0.39 & \cellcolor{gray!30}0.94 & \cellcolor{gray!30}0.59 & \cellcolor{gray!30}0.18 & \cellcolor{gray!30}\underline{0.89} & 0.29 \\
\multicolumn{1}{l|}{Gemma 2 27B} & 0.46 & 0.47 & \cellcolor{gray!30}\underline{0.66} & 0.26 & 0.61 & \cellcolor{gray!30}0.31 & \cellcolor{gray!30}0.47 & \cellcolor{gray!30}0.61 & 0.64 & \cellcolor{gray!30}0.28 & \cellcolor{gray!30}0.75 & 0.29 \\
\multicolumn{1}{l|}{LLaMA 3 70B} & 0.41 & \cellcolor{gray!30}0.62 & 0.68 & 0.23 & \cellcolor{gray!30}0.44 & \cellcolor{gray!30}0.36 & \cellcolor{gray!30}0.42 & 0.46 & 0.68 & 0.23 & 0.37 & 0.34 \\
\multicolumn{1}{l|}{Qwen 2.5 72b} & 0.37 & 0.97 & \cellcolor{gray!30}0.55 & 0.21 & 0.80 & \cellcolor{gray!30}0.31 & \cellcolor{gray!30}0.38 & \cellcolor{gray!30}0.98 & 0.53 & 0.21 & \cellcolor{gray!30}0.94 & 0.30 \\ \midrule
\multicolumn{13}{c}{Few-Shot} \\ \midrule
\multicolumn{1}{l|}{Qwen 2.5 7b} & 0.17 & \underline{0.98} & \cellcolor{gray!30}\underline{0.27} & 0.09 & \underline{0.87} & 0.21 & \cellcolor{gray!30}\underline{0.18} & \cellcolor{gray!30}\underline{0.99} & \underline{0.26} & 0.09 & \cellcolor{gray!30}\underline{0.95} & \cellcolor{gray!30}0.24 \\
\multicolumn{1}{l|}{LLaMA 3 8B} & 0.19 & \cellcolor{gray!30}\underline{0.98} & 0.28 & 0.07 & \cellcolor{gray!30}\underline{0.88} & 0.11 & \cellcolor{gray!30}0.22 & \underline{0.86} & \cellcolor{gray!30}\underline{0.31} & \cellcolor{gray!30}0.09 & \underline{0.68} & \cellcolor{gray!30}0.16 \\
\multicolumn{1}{l|}{Mistral 24B} & \cellcolor{gray!30}0.22 & 0.30 & \cellcolor{gray!30}0.33 & 0.09 & 0.41 & \cellcolor{gray!30}0.09 & 0.17 & \cellcolor{gray!30}0.47 & 0.22 & 0.09 & \cellcolor{gray!30}0.70 & 0.06 \\
\multicolumn{1}{l|}{Gemma 2 27B} & \underline{0.47} & \underline{0.98} & \cellcolor{gray!30}0.65 & \underline{0.28} & \underline{0.91} & 0.35 & 0.47 & \cellcolor{gray!30}\underline{0.99} & 0.64 & 0.28 & \underline{0.91} & \underline{0.35} \\
\multicolumn{1}{l|}{LLaMA 3 70B} & 0.42 & \underline{0.96} & 0.67 & \cellcolor{gray!30}\underline{0.25} & \cellcolor{gray!30}\underline{0.97} & \underline{0.37} & 0.42 & \cellcolor{gray!30}\underline{0.98} & 0.67 & \underline{0.24} & \underline{0.96} & 0.37 \\
\multicolumn{1}{l|}{Qwen 2.5 72b} & \underline{0.38} & 0.97 & \cellcolor{gray!30}\underline{0.57} & \underline{0.22} & \underline{0.94} & \cellcolor{gray!30}\underline{0.35} & 0.38 & \cellcolor{gray!30}\underline{0.99} & 0.56 & 0.22 & \cellcolor{gray!30}\underline{0.96} & \underline{0.33} \\ \midrule
\multicolumn{13}{c}{Chain-Of-Thought} \\ \midrule
\multicolumn{1}{l|}{Qwen 2.5 7b} & 0.17 & \cellcolor{gray!30}0.94 & 0.23 & 0.09 & \cellcolor{gray!30}0.48 & \underline{0.22} & 0.17 & 0.74 & \cellcolor{gray!30}0.24 & 0.09 & 0.36 & \cellcolor{gray!30}0.24 \\
\multicolumn{1}{l|}{LLaMA 3 8B} & \cellcolor{gray!30}\underline{0.31} & 0.32 & \cellcolor{gray!30}\underline{0.40} & \cellcolor{gray!30}\underline{0.15} & 0.17 & \cellcolor{gray!30}\underline{0.26} & 0.23 & \cellcolor{gray!30}0.39 & 0.30 & 0.09 & \cellcolor{gray!30}0.32 & \underline{0.23} \\
\multicolumn{1}{l|}{Mistral 24B} & \underline{0.39} & \cellcolor{gray!30}\underline{0.97} & 0.57 & 0.17 & \cellcolor{gray!30}\underline{0.72} & \cellcolor{gray!30}\underline{0.32} & \cellcolor{gray!30}\underline{0.41} & 0.94 & \cellcolor{gray!30}0.59 & \cellcolor{gray!30}\underline{0.20} & 0.58 & \underline{0.30} \\
\multicolumn{1}{l|}{Gemma 2 27B} & 0.46 & 0.66 & \cellcolor{gray!30}0.65 & 0.27 & 0.53 & \cellcolor{gray!30}0.35 & 0.46 & \cellcolor{gray!30}0.86 & 0.64 & 0.27 & \cellcolor{gray!30}0.80 & 0.31 \\
\multicolumn{1}{l|}{LLaMA 3 70B} & \underline{0.43} & 0.53 & \cellcolor{gray!30}\underline{0.69} & 0.23 & 0.36 & 0.36 & \cellcolor{gray!30}\underline{0.45} & \cellcolor{gray!30}0.79 & 0.68 & 0.23 & \cellcolor{gray!30}0.63 & \cellcolor{gray!30}0.37 \\
\multicolumn{1}{l|}{Qwen 2.5 72b} & 0.37 & \cellcolor{gray!30}\underline{0.98} & 0.54 & 0.21 & \cellcolor{gray!30}0.77 & \cellcolor{gray!30}0.34 & \cellcolor{gray!30}\underline{0.41} & 0.97 & \cellcolor{gray!30}\underline{0.58} & \cellcolor{gray!30}0.22 & 0.55 & 0.31 \\
\bottomrule
\end{tabular}%
}
\end{table*}

To evaluate the effectiveness of the \dataset benchmark, we conduct two sets of experiments. 
First, we use \dataset as a benchmark to assess the performance of various LLMs across different question types, categories, and subtypes. 
Second, we use \dataset as a training resource to fine-tune an LLM and investigate the value of the dataset for model adaptation.

\paragraph{Model performance.}
We evaluate multiple LLMs on the \dataset benchmark using two equivalent prompts, one in English and one in Persian. 
Tables~\ref{tbl:boolean}, \ref{tbl:multichoice}, and \ref{tbl:factoid} report the results across all configurations.

Overall, results indicate that Persian prompts consistently outperform English prompts across Boolean, Multiple-choice, and Factoid tasks. 
This is likely because the questions themselves are written in Persian, and Persian prompts better guide models to interpret them faithfully. 
Regarding prompting strategies, for Boolean and Multiple-choice questions, \textit{chain-of-thought} prompting provides the strongest performance, likely due to the reasoning-oriented nature of the questions of these tasks. 
In contrast, for Factoid questions, \textit{few-shot} prompting performs best, suggesting that Factoid questions benefit more from example-driven prompting than from stepwise reasoning, in line with prior observations~\cite{chada-natarajan-2021-fewshotqa}.

\paragraph{Fine-tuning.}
We fine-tune two LLMs---LLaMA~3~8B and its Persian variant, Dorna---on the training portion of \dataset. 
To create the train–test split, we sample 120 QA pairs from each of the 18 configuration instances, yielding 2{,}160 QA pairs for the test set. 
The remaining 8{,}640 QA pairs are used for training. 
We then evaluate both models in their vanilla and fine-tuned settings. 
For inference, based on the findings of the \textit{Model Performance} experiment, we use Persian prompts; Chain-of-Thought prompting is applied for Boolean and multiple-choice questions, and few-shot prompting is used for factoid questions. 
Figure~\ref{fig:finetune_results} presents the results.

As expected, Dorna outperforms LLaMA~3~8B, likely due to its pretraining on Persian corpora. 
Fine-tuning further yields substantial performance gains for both models, with fine-tuned Dorna achieving the strongest overall performance. 
These findings demonstrate that \dataset is valuable not only as an evaluation benchmark but also as an effective resource for adapting Persian-capable LLMs.

\section{Conclusion}\label{s:conclusion}
We presented \dataset, the first open-domain benchmark for reasoning QA in Persian. With 10{,}800 questions spanning multiple formats, reasoning categories, and difficulty levels, \dataset\ provides broad and controlled coverage of QA phenomena. Its structured taxonomy and multi-stage validation ensure high linguistic, factual, and structural quality.
Experiments show that current multilingual and Persian models still struggle with Persian reasoning tasks, while fine-tuning on \dataset\ yields notable improvements for most of the configurations. These results demonstrate the benchmark’s effectiveness for both evaluation and model adaptation.

\dataset\ fills a major resource gap in Persian NLP and IR, establishing a foundation for future work in multilingual reasoning and LLM development. As future directions, we aim to explore retrieval-augmented generation (RAG) and more advanced reasoning-oriented models such as DeepSeek~\cite{guo2025deepseek}.

\bibliographystyle{ACM-Reference-Format}
\bibliography{Main}

\end{document}